\documentclass[journal,compsoc]{IEEEtran}
\usepackage{algorithm,algorithmic}
\usepackage{multirow}
\usepackage[table]{xcolor}

\usepackage{amsmath}
\usepackage{hyperref}
\hypersetup{colorlinks = true,
            linkcolor = blue,
            urlcolor = blue, 
            citecolor=red}
\renewcommand\footnoterule{\kern-3pt \hrule width 3.5in \kern 2.6pt}
\newcommand\Tstrut{\rule{0pt}{2.3ex}}       
\makeatletter
\def\BState{\State\hskip-\ALG@thistlm}
\makeatother

\ifCLASSINFOpdf
  \usepackage[pdftex]{graphicx}
\else
\fi
\ifCLASSOPTIONcompsoc
\usepackage[caption=false,font=normalsize,labelfont=sf,textfont=sf]{subfig}
\else
  \usepackage[caption=false,font=footnotesize]{subfig}
\fi
\begin{document}
\title{$\lambda$-Scaled-Attention: A Novel Fast Attention Mechanism for Efficient Modeling of Protein Sequences}

\author{Ashish~Ranjan,
		Md Shah Fahad, and
        Akshay Deepak
    
\IEEEcompsocitemizethanks{
	\IEEEcompsocthanksitem Ashish Ranjan (corresponding author), Md Shah Fahad and Akshay Deepak are with the Department of Computer Science \& Engineering, National Institute of Technology Patna, India. E-mail: \{ashish.cse16, shah.cse16, akshayd\}@nitp.ac.in.}
}

%

\IEEEtitleabstractindextext{
\begin{abstract}
Attention-based deep networks have been successfully applied on textual data in the field of NLP. However, their application on protein sequences poses additional challenges due to the weak semantics of the protein words, unlike the plain text words. These unexplored challenges faced by the standard attention technique include (i) vanishing attention score problem and (ii) high variations in the attention distribution. In this regard, we introduce a novel $\lambda$-\textit{scaled attention} technique for fast and efficient modeling of the protein sequences that addresses both the above problems. This is used to develop the $\lambda$-\textit{scaled attention} network and is evaluated for the task of protein function prediction implemented at the protein sub-sequence level. Experiments on  the datasets for biological process (BP) and molecular function (MF) showed significant improvements in the F1 score values for the proposed $\lambda$-\textit{scaled attention} technique over its counterpart approach based on the standard attention technique (+2.01\% for BP and +4.67\% for MF) and state-of-the-art ProtVecGen-Plus approach (+2.61\% for BP and +4.20\% for MF). Further, fast convergence (converging in half the number of epochs) and efficient learning (in terms of very low difference between the training and validation losses) were also observed during the training process.
\end{abstract}

}

\maketitle

\IEEEpeerreviewmaketitle

\section{Introduction}
\IEEEPARstart{A}{utomatic} functional characterization of proteins is one of the key challenges in computational biology. The proteins are fundamental to the biological interactions. They are found helpful in, among other things, studying the attack of viruses \cite{spike_protein1, spike_protein2, bacteria} and harmful bacteria \cite{bacteria} on the organism. Proteins, in their elementary form, are linear chains of amino acids called ``protein sequences", commonly also referred to as ``the language of life" \cite{lang_life}. In the past few years, undoubtedly, protein sequences have become a cheap and reliable source for the protein studies. The surplus amount of protein sequences -- generated due to the blooming of sequencing technologies \cite{NextGenSeq} -- has marked a transitional shift toward the data-driven  machine learning approaches \cite{FANNGO, MLDA_protFun} for the characterization of proteins.

The evolution of deep-learning has further taken the automated characterization process to the next level by significantly bringing down the effort needed to process protein sequences. In particular, the deep recurrent neural networks (DRNN), e.g., long-short term memory (LSTM) networks, have shown great potential with protein sequences for the function prediction task \cite{Paper_mine, DNA_bind, anticancer}. In protein sequences, which are also represented as a string of \textit{n}-mers \cite{Paper_mine, prolango}, the order of amino acids/\textit{n}-mers has a significant effect in determining their function(s). These orderly arrangements are believed to create orderly dependencies/patterns proven useful in determining their function(s). DRNN helps capture such orderly dependencies between the amino acids/\textit{n}-mers of a protein sequence.

However, it should be noted that protein words (either amino acid or \textit{n}-mers) that make up a protein sequence have a different significance in uncovering the underlying \textit{functional context} in the sequence. The set of key significant protein words, when combined, further provides a clear and strong indication of the \textit{functional context}. The \textit{functional context} refers to the ``small pattern" common among the protein sequences of the same functional family. These are created due to the orderly dependencies between the sub-set of protein words making up a protein sequence, e.g.,  \textit{sequence motifs} \cite{Seq_motif}. 
Thus, there are two important aspects to efficient processing of protein sequences: (i) learning the orderly dependencies between the protein words, and (ii) capturing the importance of individual protein words toward revealing the underlying \textit{functional context}.

While LSTM can capture the orderly dependencies between the protein words, it has no mechanism to highlight the significance of words. This causes masking/losing the critical information (possibly, containing the \textit{functional context}), resulting in a sub-optimal intermediate representation. Therefore, the \textit{attention} mechanism \cite{Attention} that builds an intermediate representation by prioritizing the key informative words, regardless of their position in the sentence, will be useful in such cases and has been explored in this work. \textit{Attention}-based DRNN has already become a standard solution for a wide variety of tasks, such as  document classification \cite{hierarchical_attention} and sentiment classification \cite{attention_sentiment1, attention_sentiment2}. They have a profound effect on the produced intermediate representations.

\begin{figure}[!t]
	\centering
	\includegraphics[width=3.5in]{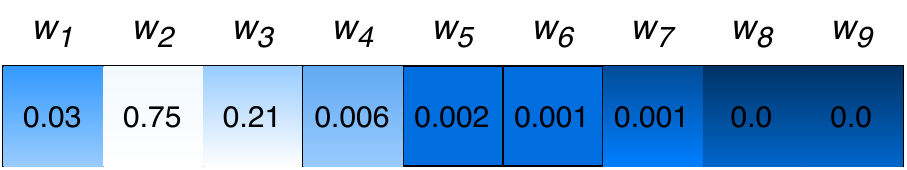}
	\caption{Case I: Distribution of attention scores with only a small number of significant words \{w$_2$, w$_3$\}.}
	\label{attn1}
\end{figure}

\begin{figure}[!t]
	\centering
	\includegraphics[width=3.5in]{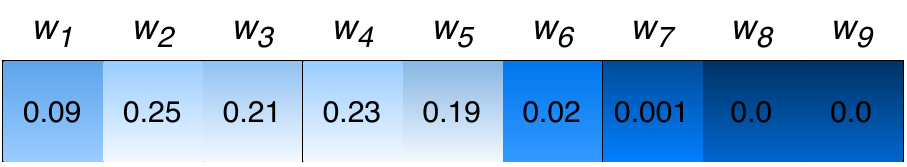}
	\caption{Case II: Distribution of attention scores with a relatively large number of significant words \{w$_2$, w$_3$, w$_4$, w$_5$\}.}
	\label{attn2}
\end{figure}

Howsoever, the application of \textit{attention} technique on protein sequences has its own set of challenges. This is because, unlike the plain text words which have strong \textit{context-specific} semantics, protein words (i.e., amino acid/\textit{n}-mers) have a very weak \textit{context-specific} semantics. Thus, while a small sub-set of words is sufficient to identify the context of a plain text sentence, possibly a large sub-set of protein words is needed to identify a \textit{functional context} in the protein sequence. The large sub-set of protein words helps to make a strong case for the underlying \textit{functional context}. The challenges faced by the \textit{attention} technique are as follows:
\begin{itemize}
	\item \textbf{Vanishing attention score problem}: \textit{Attention} \cite{Attention} technique, in general, produces a distribution that helps identify the significance of words in a sentence. While for the plain text sentences, standard \textit{attention} \cite{Attention} performs reasonably well with only a few significant words to focus, it struggle to reproduce the same performance for the protein sequences. The reason for this is the vanishing of the attention scores caused due to large sub-set of significant protein words that form a \textit{functional context}.
	
	\hspace{4.5mm} For example, consider the distribution of attention scores as shown in Figures \ref{attn1} and \ref{attn2}. In Figure \ref{attn1}, where there are only a few significant words in the sentence, the attention score is high (for $w_2 = 0.75$). However, when the number of significant words in a sentence is increased, the attention scores become small (see Figure \ref{attn2}). A further increase in the number of significant words will soon start to produce diminished attention scores. 
	
	\hspace{4.5mm} In the latter case (Figure \ref{attn2}), when the attention scores become too small, it may cause a restricted flow of information down the network. We call this problem, the ``\textit{vanishing attention score}'' problem.
	
	\item \textbf{High variations in the attention distributions}: The attention score distributions for different protein sequences, in general, varies significantly. This creates problem capturing the high variations among the allocated scores, making the training process less efficient.
\end{itemize}

To address the above discussed issues, a novel \textit{scaled-attention} technique that produces attention scores via the scale-up operation is proposed in this work that helps eliminate the ``\textit{vanishing attention score problem}", irrespective of the number of significant protein words in the protein sequence. The proposed solution also helps in reducing the high variation in the attention distribution. In light of the above discussion, the major contributions of the paper are:
\begin{itemize}
	\item A novel $\lambda$-\textit{scaled attention} technique for the fast and efficient modeling of protein sequences. The $\lambda$ parameter is useful in controlling the convergence of model while training. A high value of $\lambda$ will push for fast convergence, while a low value of $\lambda$ will cause slow convergence.
	
	\item The proposed $\lambda$-\textit{scaled attention} technique is next used to develop a general deep neural network architecture for the protein sequence, called the ``$\lambda$-\textit{scaled attention} network", which is evaluated for the protein function prediction task implemented at the protein sub-sequence level.
	
	\item Experiments on the datasets for biological process (BP) and molecular function (MF) demonstrate that the proposed $\lambda$-\textit{scaled attention} technique has a major impact on improving the overall performance. Compared to its counterpart approach based on standard attention \cite{Attention} with segment size = 100, it shows improvements in F1-score by a margin of +2.01\% (BP) and +4.67\% (MF), respectively. The corresponding improvements over state-of-the-art multi-segment based ProtVecGen-Plus approach \cite{Paper_mine} are +2.61\% (BP) and +4.20\% (MF), respectively. 
	
	\item Further, during training, the proposed $\lambda$-\textit{scaled attention} technique converges faster (taking around half the amount of time taken by the baseline methods) and is efficient (in terms of very low difference between the training and validation losses) when compared to both the simple \textit{baseline} and \textit{baseline+attention} based classifiers (see Sec. \ref{fastnefficient} for more details). 
\end{itemize}

The organization of rest of the paper is as follows: Section \ref{sec_model} outlines the proposed $\lambda$-\textit{Scaled Attention Network}, following which Section \ref{Prot_fun_pred} discuss the use of the $\lambda$-\textit{Scaled Attention Network} for protein function prediction. This follows results and discussion in Section \ref{results}. Finally, Section \ref{conclusion} concludes the paper.

\section{\textbf{$\lambda$-\textit{Scaled Attention Network}}}\label{sec_model}
In this section, we introduce a novel $\lambda$-\textit{scaled attention network} that can be applied efficiently for a wide range of classification tasks using protein sequences. However, in this paper the proposed network has been evaluated for the  task of protein function prediction. The proposed architecture is composed of the following components: (1) Protein word embedding layer, (2) Deep recurrent processing layer, (3) Protein word-level $\lambda$-scaled attention layer, and (4) Dense output layer. Moreover, we also added dropout layers \cite{dropout} before the deep recurrent processing and dense output layers, which helps avoid the over-fitting problem. The corresponding amount of disconnections with each dropout layers were taken as 0.3\% and 0.2\%, respectively. These are described in Sections \ref{sec_embdd} -- \ref{sec_dense}. Before their description,  the preliminary notations and protein sequence tokenization (as the pre-processing step) are discussed in Sections \ref{sec_notation} -- \ref{sec_token}.

\subsection{Notations}\label{sec_notation}
Let $\{GO_1, GO_2,...,GO_K\}$ denote the set of `$K$' $GO$-terms (either for biological process or molecular function). Let $ P = \{(P_1, Y_1), (P_2,Y_2), ..., (P_m,Y_m)\} $ denote a database of $m$ labeled protein sequences, where $P_i$ denotes a protein sequence and $Y_i$ = $\{y_{i1}, y_{i2},..., y_{ik}...,y_{iK}\}$ denotes one-hot encoding representing the set of $GO$-term annotations corresponding to the protein sequence $P_i$, such that $y_{ik}=1$  if $P_i$ exhibits $GO_{k}$ and 0 otherwise.

\subsection{Protein Sequence Tokenization}\label{sec_token} 
The tokenization of protein sequences involves decomposing protein sequences into a string of amino acids or \textit{n}-mers, which are assumed to be the protein words for the protein sequences. Since the length of  protein sequences is highly variable in nature, for simplification, we assume the maximum permissible length for protein sequences to be \textit{max\_seq\_len}. Protein sequences with length $>$ \textit{max\_seq\_len} are truncated, while padding is done for shorter protein sequences. The maximum number of possible protein words in a sequence is:
\begin{equation}\label{max_seq_len}
T = \textit{max\_seq\_len} - n + 1
\end{equation}
where, \textit{n} is the size of \textit{n}-mers / protein words. 

\subsection{Protein Word Embedding Layer}\label{sec_embdd}
Tokenized protein sequences are passed through the embedding layer that helps in creating meaningful dense representations for the protein words in the sequence. The output from the embedding layer is a matrix of size [$T$ x \textit{d}], where $T$ denotes the maximum number of possible protein words in a sequence (Equation \ref{max_seq_len}) and \textit{d} denotes the embedding dimension.

\subsection{Deep Recurrent Processing Layer}\label{sec_recc}
To accomplish the learning of the orderly dependencies between the protein words, deep recurrent neural networks (DRNN), such as GRU \cite{GRU} / LSTM \cite{LSTM} network, are used to process protein sequences. These networks are linear chain of recurrent units (e.g., GRU cell / LSTM cell) that process sequences, word-by-word, and produce a hidden state vector ($h_t$) corresponding to every protein word. 

Let at any time-step (say $t$), $x_t$ represent the current input and $h_{t-1}$ represent the previous hidden state. Then, the recurrent unit accepts $x_t$ and $h_{t-1}$ as the input and produces $h_{t}$ as the corresponding output. Note that $x_t$ actually refers to the protein word embedding occurring at time-step $t$. A brief description of both GRU and LSTM cells are given below.

\textit{a) Gated Recurrent Unit (GRU).}
The major components with the GRU cells are:  \textit{update gate} and \textit{reset gate}. The \textit{update gate} helps to measure the usefulness of the past information ($h_{t-1}$) for the current time-step $t$. The output of the \textit{update gate}, denoted $z_t$, is described in Eq. \ref{update_logic}. The \textit{reset gate}, on the other hand, helps to forget the past information that may be irrelevant in the future. The output of the \textit{reset gate}, denoted $r_t$, is described in Eq. \ref{reset_logic}.
\begin{equation}\label{update_logic}
z_t = \sigma(W_zx_t + U_zh_{t-1} + b_z)
\end{equation}
\begin{equation}\label{reset_logic}
r_t = \sigma(W_rx_t + U_rh_{t-1} + b_r)
\end{equation}
where,\\
-- $W_z$, $W_r$, $U_z$, and $U_r$ denote weight matrices,\\ 
-- $b_z$ and $b_r$ denote bias vectors.

The outputs $z_t$ and $r_t$ help generate the hidden state vector $h_t$ as follows:
\begin{equation}\label{final_memory}
h_t = [z_t \odot h_{t-1}] + [(1 - z_t) \odot \tilde{h_t}]
\end{equation}
where, $\tilde{h_{t}}$ denotes the current memory content and is computed as follows:
\begin{equation}\label{current_memory}
\tilde{h_{t}} = \tanh(W_hx_t + (r_t \odot U_hh_{t-1}) + b_h)
\end{equation}
where,\\
-- $W_h$ and $U_h$ denote weight matrices,\\
-- $b_h$ denotes the bias vector, and\\
-- $\odot$ denotes the element-wise multiplication.
 
\begin{figure}[!t]
	\centering
	\includegraphics[width=3.5in]{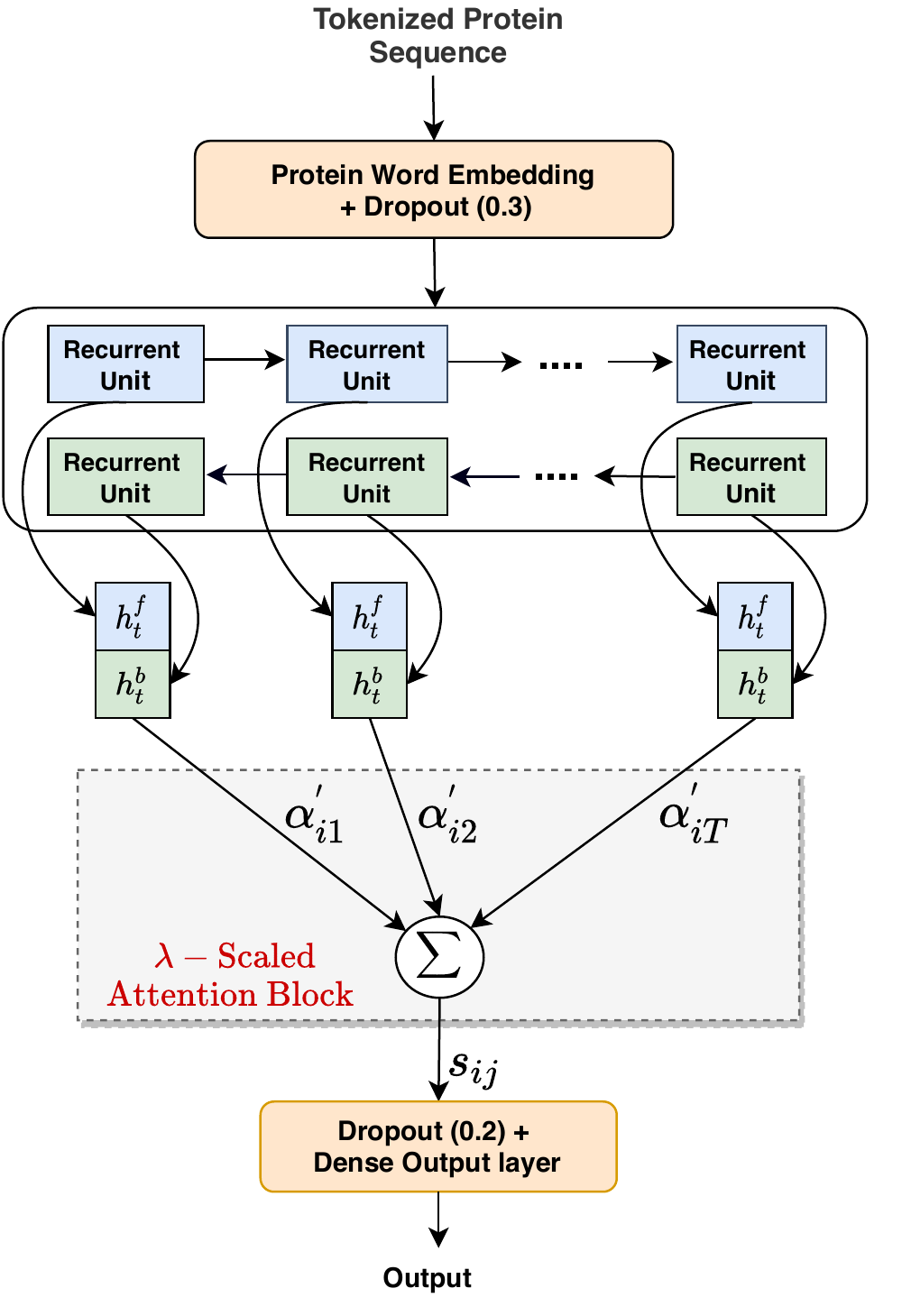}
	\caption{$\lambda$-\textit{Scaled Attention Network}: $\lambda$-scaled-attention based deep recurrent network for the protein sequences.}
	\label{SegVecGenerator}
\end{figure}

\textit{b) Long Short-Term Memory (LSTM).}
The key component of an LSTM cell is the memory element called \textit{cell state} that maintains the global information. Other major components with an LSTM cell include: \textit{forget gate} and \textit{input gate}, which assist \textit{cell state} in maintaining the global information. The \textit{forget gate} assists \textit{cell state} to forget the irrelevant past information. The output of the \textit{forget gate}, denoted as $f_t$, is described in Eq. \ref{forget_logic}. The \textit{input gate} assists in adding new information to the \textit{cell state}. The outputs of the \textit{input gate}, denoted as $i_t$ and $\bar{C_t}$, are described in Eq. \ref{input_logic}.
\begin{equation}\label{forget_logic}
f_t = \sigma(W_f[h_{t-1},x_t] + b_f)
\end{equation}

\begin{equation}\label{input_logic}
\begin{aligned}
i_t = \sigma(W_i[h_{t-1},x_t] + b_i)\\ 
\bar{C_t} = tanh(W_c[h_{t-1},x_t] + b_c)
\end{aligned}
\end{equation}
where,\\
-- $W_f$, $W_i$, and $W_c$ denotes weight matrices,\\
-- $b_f$, $b_i$, and $b_c$ denote bias vectors.

The \textit{cell state} is updated as follows:
\begin{equation}\label{cell_state_update}
C_t = (f_t \odot C_{t-1}) + (i_t \odot \bar{C_t})
\end{equation}
where, $\odot$ represents element-wise multiplication.

The hidden state $h_t$ is obtained based on the dot product between the updated \textit{cell state} $C_t$ and the \textit{output} logic $o_t$. They are described as:

\begin{equation}
o_t = \sigma(W_o[h_{t-1},x_t] + b_o)
\end{equation}
\begin{equation}
h_t = o_t \odot tanh(C_t)
\end{equation}
where, $W_o$ and $b_o$ are the weight matrix and bias vector respectively, with the \textit{output} gate.

\textbf{\textit{Bi-directional recurrent processing}}:
In this work, the deep recurrent processing layer is implemented with the bi-directional configuration \cite{BiLstm} comprising of forward and backward DRNN's. The bi-directional property allows computing the hidden state for the protein words, which is based on both the preceding as well as the following words in a protein sequence. Let $[x_1, x_2, ..., x_t, ...,x_T]$ represent the ordered protein words in a protein sequence, where $ T $ is the total number of time-steps equivalent to the maximum possible protein words in a sequence (Eq. \ref{max_seq_len}).

The forward DRNN reads the protein sequences starting from $x_1$ to $x_T$ and the backward DRNN reads the protein sequences backwards from $x_T$ to $x_1$. Let $h_t^f$ and $h_t^b$ represent the hidden states for the word $x_t$ obtained from the forward and backward DRNNs, respectively. The final hidden state for the protein word $x_t$ is the concatenation of $h_t^f$ and $h_t^b$, i.e., $h_i$ = [$h_t^f$, $h_t^b$]. The dimension of the hidden states from both the forward and backward DRNN is taken as 70.

\subsection{$\lambda$-Scaled Attention Layer}\label{sec_attn}
Here we introduce a novel $\lambda$-\textit{scaled attention} technique to address both the \textit{vanishing attention score} problem and the variations in the attention score distribution. The proposed attention technique allows to find a good representation of the protein sequences by aggregating the weighted representation of their protein words.

Let $[h_{i1}, h_{i2}, ..., h_{it}, ..., h_{iT}]$ represent the list of hidden state vectors for the $i^{th}$ protein sequence, where $h_{it}$ is the hidden state vector corresponding to the protein word occurring at time $t$. Formally, the proposed $\lambda$-scaled attention technique can be defined as consisting of following steps:
\begin{enumerate}
\item Each hidden state vector $h_{it}$, where $t \in \{1,2,...,T\}$, is fed forward through a single layer perceptron to obtain the corresponding hidden representation $u_{it}$ as:
\begin{equation}
u_{it} = tanh(W_ah_{it} + b_a)
\end{equation}
where, $W_a$ and $b_a$ represent the weight matrix and bias element, respectively.

\item Next, the scaled attention score $\alpha_{it}^{'}$  for the protein words occurring at time-steps $t \in \{1,2,...,T\}$ is calculated as follows:
\begin{equation}\label{scaled-softmax2}
\alpha_{it}^{'} = \frac{\alpha_{it}}{\alpha_{it}^{max}}; \hspace{6mm} \text{s.t.} \hspace{2mm} \alpha_{it}^{'} \leq 1
\end{equation}
 where,
\begin{equation}\label{softmax}
\alpha_{it} = \frac{exp(u_{it}^T)}{\sum_{t=1}^{T}exp(u_{it}^T)}
\end{equation}

\begin{equation}\label{scaled-softmax1}
\alpha_{it}^{max} = max[\alpha_{i1}, \alpha_{i2}, ..., \alpha_{it}, ..., \alpha_{iT}]
\end{equation}

Equation \ref{softmax} represents the standard \textit{softmax} operation, which is used to find the distribution of attention scores $\alpha_{it}, \text{where } t \in \{1,2,...,T\}$. Following this in Eq. \ref{scaled-softmax1}, the value of the maximum attention score $\alpha_{it}^{max}$ is computed from the distribution. These are used in  Eq. \ref{scaled-softmax2} to compute the scaled-up attention scores; the maximum possible scaled-up attention score is  1.

\hspace{4.5mm} The scaling operation in Eq. \ref{scaled-softmax2} creates the amplified attention scores, thereby, eliminating the \textit{vanishing attention score problem}. Moreover, scaling also ensures that attention scores for each time-step are now on the same scale, which reduces the variations in the attention score distribution.

\item Optionally, a scalar parameter ``$\lambda$'' can also be multiplied to the scaled attention scores:  $[\alpha_{i1}^{'}, \alpha_{i2}^{'}, ..., \alpha_{it}^{'}, ..., \alpha_{iT}^{'}]$, which limits the maximum value a scaled attention score can take. Such restriction on the maximum possible scaled attention score is useful in controlling the training rate. A higher value of ``$\lambda$" will push for faster training, while a lower value of ``$\lambda$" will slow-down the training rate.

\item The final vector $s_i$ for the $i^{th}$ protein sequence is the aggregation of the weighted hidden states. The weighted hidden states are computed by multiplying the scaled attention scores $\alpha_{it}^{'}$s with the corresponding hidden states $h_{it}$s:
\begin{equation}\label{mul_attn}
s_i = \sum_{t=1}^{T}\alpha_{it}^{'}h_{it}
\end{equation}
\end{enumerate}
The benefits of the proposed $\lambda$-scaled attention are as follows:
\begin{itemize}
	\item It helps overcome the \textit{vanishing attention score} problem, which otherwise causes restricted flow of the information.
	
	\item It further helps in reducing the high variations in the attention distribution, making the training procedure a more efficient one.
\end{itemize}

\subsection{Dense Output Layer}\label{sec_dense}
Following the attention layer is a fully-connected dense output layer, where the number of neurons is equal to the number of \textit{GO}-terms (i.e., \textit{K}). The output neurons are implemented with the \textit{sigmoid} activations in order to deal with the multi-label scenario. Note that the protein function prediction is a multi-label classification problem. Other parameters include, ``\textit{binary-cross entropy}" as the loss function and ``\textit{adam}" \cite{adam} as the gradient optimizer. 

\section{Protein Function Prediction}\label{Prot_fun_pred}
In this section, we discuss how we utilize the proposed $\lambda$-\textit{scaled attention} network for the task of protein function prediction. We choose to use the $\lambda$-\textit{scaled attention} network at the protein sub-sequence level for classifying protein sub-sequences. This is because, in general, protein words forming up the \textit{functional context(s)} within the protein sequences do not share long-distance dependency. The advantage of using protein sub-sequences is also discussed in our earlier work \cite{Paper_mine}. The classified protein sub-sequences are eventually used to determine the function(s) of their parent sequence. 

The proposed solution has the following steps: 1) \textit{Global Protein Sequence Representation}: an approach to generate global representation for protein sequences based on small segments of protein sequences and the proposed $\lambda$-\textit{scaled attention} network, and 2) \textit{Classification Model} : a multi-layer perceptron (MLP) network based classifier to predict the  function(s) of unlabeled protein sequences. These are described next.

\subsection{\textbf{Global Protein Sequence Representation}}\label{global_seq_rep}
This is a two-step procedure: (1) Training a $\lambda$-\textit{scaled attention} network using the protein segments, and (2) Utilizing the trained $\lambda$-\textit{scaled attention} network for producing the global representation for the protein sequences.

\subsubsection{Training a $\lambda$-\textit{scaled attention} network using the protein segments}\label{pre_train}
\begin{figure}[!t]
	\centering
	\includegraphics[width=3.4in]{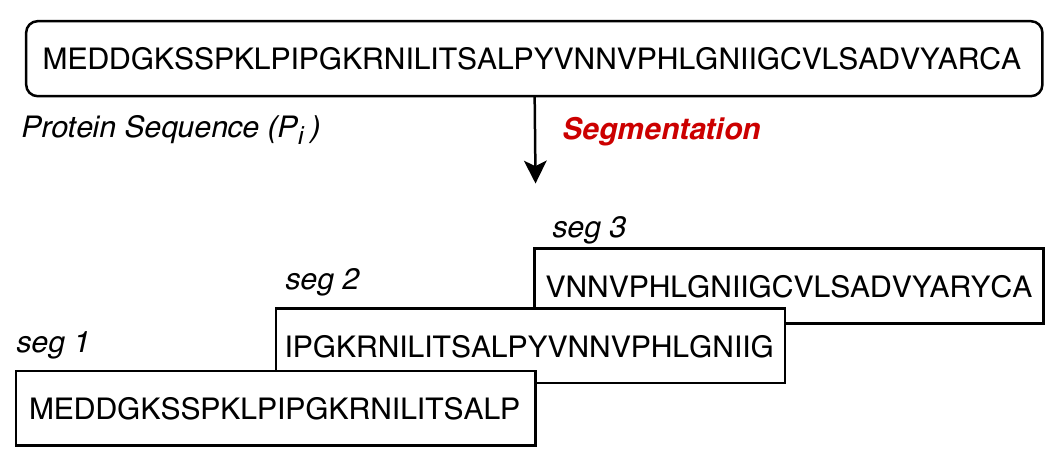}
	\caption{Protein sequence segmentation.}
	\label{module_seg}
\end{figure}

The segmented dataset is used to train the proposed $\lambda$-\textit{scaled attention} network, which is constructed from the training dataset $P = \{P_i, Y_i\}_{i=1}^m$ as follows:
\begin{itemize}
	\item  As shown in Figure \ref{module_seg}, each protein sequence $P_i$ with the label-set $Y_i$ is decomposed into equal-sized protein segments. The set of such protein segments corresponding to protein $P_i$ is denoted as $\phi_i = \{p_{i,1}, p_{i,2},...,p_{i,j},...\}$, where $p_{i,j}$ represents the $j^{th}$ segment of protein $P_i$. Further, the decomposition happens in an overlapping manner. Padding is done to the last protein segment if it is smaller than the segment size.
	
	\item Any two adjacent segments, say $p_{i,j}$ and $p_{i,j+1}$, are chosen such that they have at least 50\% of region in common. 
	
	\item Each segment $p_{i,j} \in \phi_i$ is assigned the same label-set $Y_i$ as the parent protein sequence $P_i$. This constitutes the new segmented training dataset as $(p_{i,j}, Y_i)$, where $i \in \{1,2,..,m\}$ and $j \in \{1,2,...\}$.
\end{itemize}
Next, the newly constructed segmented dataset is used for the training of the $\lambda$-\textit{scaled attention} network as already discussed in Section \ref{sec_model}.

\subsubsection{Utilizing the trained $\lambda$-\textit{scaled attention} network for producing the representations for the protein sequences}
The trained $\lambda$-\textit{scaled attention} network based segment classifier from previous Section \ref{pre_train} is used to produce the global representation for protein sequences in both the training and testing datasets. This involves the following steps:
\begin{itemize}
	\item The protein sequences are decomposed into protein segments as discussed before in the Section \ref{pre_train}. Let $\phi_i = \{p_{i,1}, p_{i,2},...\}$ represent the segments for the $i^{th}$ protein sequence.
	
	\item The decomposed protein segments from the set $\phi_i$ are classified using the pre-trained segment classifier. An output from the segment classifier is a $K$-dimensional vector that represents the corresponding protein segment  $\textbf{p}_{i,j}$.
	
	\item Next, the average of segment representations is computed to formulate the intermediate sequence representation $\textbf{P}_i^{'}$ as:
	\begin{equation}
		\textbf{P}_i^{'} \longleftarrow \frac{1}{|\phi_i|}\sum_{p_{i,j} \in \phi_i} \textbf{p}_{i,j} \hspace{6mm}\textit{\# Averaging}
	\end{equation}
	For the large number of long protein sequences, only a few segments are found supportive to the true output classes. The average operation, in general, may cause lowering of the support for the true output classes. Thus, the values of $\textbf{P}_i^{'}$ are scaled-up as follows:
	\begin{equation}\label{scaling}
	\textbf{P}_i \longleftarrow \frac{\textbf{P}_i^{'}}{max(\textbf{P}_i^{'})} \hspace{9mm}\textit{\# Scaling}
	\end{equation}
\end{itemize}

\subsection{Classification Model}\label{class_model}
The problem of protein function prediction was modeled as a multi-label classification problem using the multi-layer perceptron (MLP) network. The MLP network consists of a single hidden layer followed by the output layer. The output neurons are implemented with the \textit{sigmoid} activations in order to deal with the multi-label scenario. The \textit{K}-dimensional feature representations for the protein sequences, as obtained in Section \ref{global_seq_rep}, are used for the training of the perceptron model. Other parameters include ``\textit{binary-cross entropy}" as the loss function and ``\textit{adam}" as the gradient optimizer.

\section{Result and Discussion}\label{results}
This section presents the findings of the proposed system and discusses them in depth. The proposed network architecture was implemented using Keras 2.0.6\footnote{https://github.com/keras-team/keras} with a TensorFlow backend. 

\subsection{Datasets}
For evaluation purposes, we utilized the protein sequences from the UniProtKB database \cite{uniprot}, where Gene Ontology (GO) \cite{GO} based terminology is used to indicate the output labels (i.e., protein functions) for the protein sequences. Protein sequence datasets with GO-terms defined across two different domains, Molecular Function (MF) and Biological Process (BP), were used for experiments. These two datasets are the same as the ones used in \cite{Paper_mine}.

The BP dataset consists of 58310 protein sequences (with 295 GO-terms for the biological process), while the MF dataset comprises of 43218 protein sequences (with 135 GO-terms for the molecular function). In both the datasets, the number of protein sequences corresponding to each GO-term is $\geq$ 200. The datasets are available for download at \url{https://bit.ly/2RMsyOV}. For all the experiments, the datasets were split into the train and test datasets with 75\% being in the training dataset, while the remaining 25\% belonging to the testing dataset.

\subsection{Metrics}
The proposed solution was evaluated using metrics such as average precision, average recall, and average F1-score \cite{Metrices1, Metrices2}; each metric being computed by taking the average over the corresponding performances of each individual test sample in the test dataset, with $n$ test samples. Let $Y_i$ be the set of true $GO$-terms and $Y_i^{'}$ be the set of predicted $GO$-terms. Then, the used metrics are defined as:

\begin{itemize}
	\item Average Precision: 
	\begin{equation} \label{precision}
	\text{\textit{Avg. Precision}} = \frac{1}{n}\sum_{i=1}^{n}\frac{\arrowvert Y_i \bigcap Y_i^{'} \arrowvert}{\arrowvert Y_i^{'} \arrowvert}
	\end{equation}
	
	\item  Average Recall: 
	\begin{equation} \label{recall}
	\text{\textit{Avg. Recall}} = \frac{1}{n}\sum_{i=1}^{n}\frac{\arrowvert Y_i \bigcap Y_i^{'} \arrowvert}{\arrowvert Y_i^{'} \arrowvert}
	\end{equation}
	
	\item Average F1-Score:
	\begin{equation} \label{avg_f11}
	\text{\textit{Avg. F1-score}} = \frac{1}{n}\sum_{i=1}^{n}\frac{2\arrowvert Y_i \bigcap Y_i^{'} \arrowvert}{\arrowvert Y_i\arrowvert + \arrowvert Y_i^{'} \arrowvert}
	\end{equation}
\end{itemize}

\begin{table}[t]
	\caption{Classification Accuracy with LSTM (Base. = Baseline, Attn. = Attention)}\label{LSTM_atten}
	\centering
		\begin{tabular}{|p{0.45cm}|p{0.7cm}|p{0.75cm}|p{0.7cm}|p{0.7cm}|p{0.7cm}|p{0.7cm}|p{0.7cm}|}
		\hline 
		& \multicolumn{2}{c|}{ } & \multicolumn{5}{c|}{$\lambda$-Scaled Attention Network}\Tstrut\\ [0.4ex]
		
		\hline \Tstrut
		Seg. size & Base. & Base.+ Attn. & $\lambda$ = 1.0  & $\lambda$ = 0.7 & $\lambda$ = 0.5 & $\lambda$ = 0.3 & $\lambda$ = 0.1\Tstrut\\
		\hline
		\multicolumn{8}{|c|}{Dataset : \textit{Biological Process}}\Tstrut\\ [0.4ex]
		\hline
		80 & 53.71 & 55.57 & 56.83 & 56.77 & 56.69 & 56.86 & 57.27\Tstrut\\ [0.4ex]
		100 & 52.33 & 54.95 & 56.96 & 56.85 & 57.09 & 57.01 & 56.96\Tstrut\\ [0.4ex]
		120 & 51.51 & 55.41 & 57.15 & 57.04 & 57.16 & 57.04 & 56.39\Tstrut\\ [0.4ex] \hline
		\multicolumn{8}{|c|}{Dataset : \textit{Molecular Function}}\Tstrut\\ [0.4ex]
		\hline
		80 & 66.13 & 67.41 & 69.89 & 69.86 & 69.57 & 69.02 & 69.01\Tstrut\\ [0.4ex]
		100 & 64.41 & 65.25 & 69.92 & 69.77 & 69.17 & 68.99 & 68.01\Tstrut\\ [0.4ex]
		120 & 62.26 & 65.93 & 69.80 & 69.67 & 69.29 & 68.93 & 67.32\Tstrut\\ [0.4ex] \hline
	\end{tabular}
\end{table}

\begin{table}[t]
	\caption{Classification Accuracy with GRU (Base. = Baseline, Attn. = Attention)}\label{GRU_atten}
	\centering
	\begin{tabular}{|p{0.45cm}|p{0.7cm}|p{0.75cm}|p{0.7cm}|p{0.7cm}|p{0.7cm}|p{0.7cm}|p{0.7cm}|}
		\hline 
		 & \multicolumn{2}{c|}{} & \multicolumn{5}{c|}{$\lambda$-Scaled Attention Network}\Tstrut\\ [0.4ex]
		
		\hline \Tstrut
		Seg. size & Base. & Base.+ Attn. & $\lambda$ = 1.0  & $\lambda$ = 0.7 & $\lambda$ = 0.5 & $\lambda$ = 0.3 & $\lambda$ = 0.1\Tstrut\\
		\hline
		\multicolumn{8}{|c|}{Dataset : \textit{Biological Process}}\Tstrut\\ [0.4ex]
		\hline
		80 & 54.21 & 56.48 & 57.43 & 57.14 & 57.09 & 56.94 & 56.35\Tstrut\\ [0.4ex]
		100 & 52.66 & 55.60 & 57.28 & 57.16 & 57.11 & 57.00 & 56.28\Tstrut\\ [0.4ex]
		120 & 51.95 & 55.28 & 57.34 & 57.33 & 57.28 & 56.87 & 56.33\Tstrut\\ [0.4ex] \hline
		\multicolumn{8}{|c|}{Dataset : \textit{Molecular Function}}\Tstrut\\ [0.4ex]
		\hline
		80 & 66.72 & 66.67 & 69.99 & 70.09 & 69.58 & 69.33 & 67.70\Tstrut\\ [0.4ex]
		100 & 61.75 & 64.95 & 70.22 & 69.81 & 69.72 & 69.09 & 66.73\Tstrut\\ [0.4ex]
		120 & 61.59 & 64.96 & 69.69 & 69.68 & 69.53 & 69.25 & 67.81\Tstrut\\ [0.4ex] \hline
	\end{tabular}
\end{table}

\begin{figure*}[!t]
	\centering
	\subfloat[Training Loss]
	{\includegraphics[width=9.1cm]
		{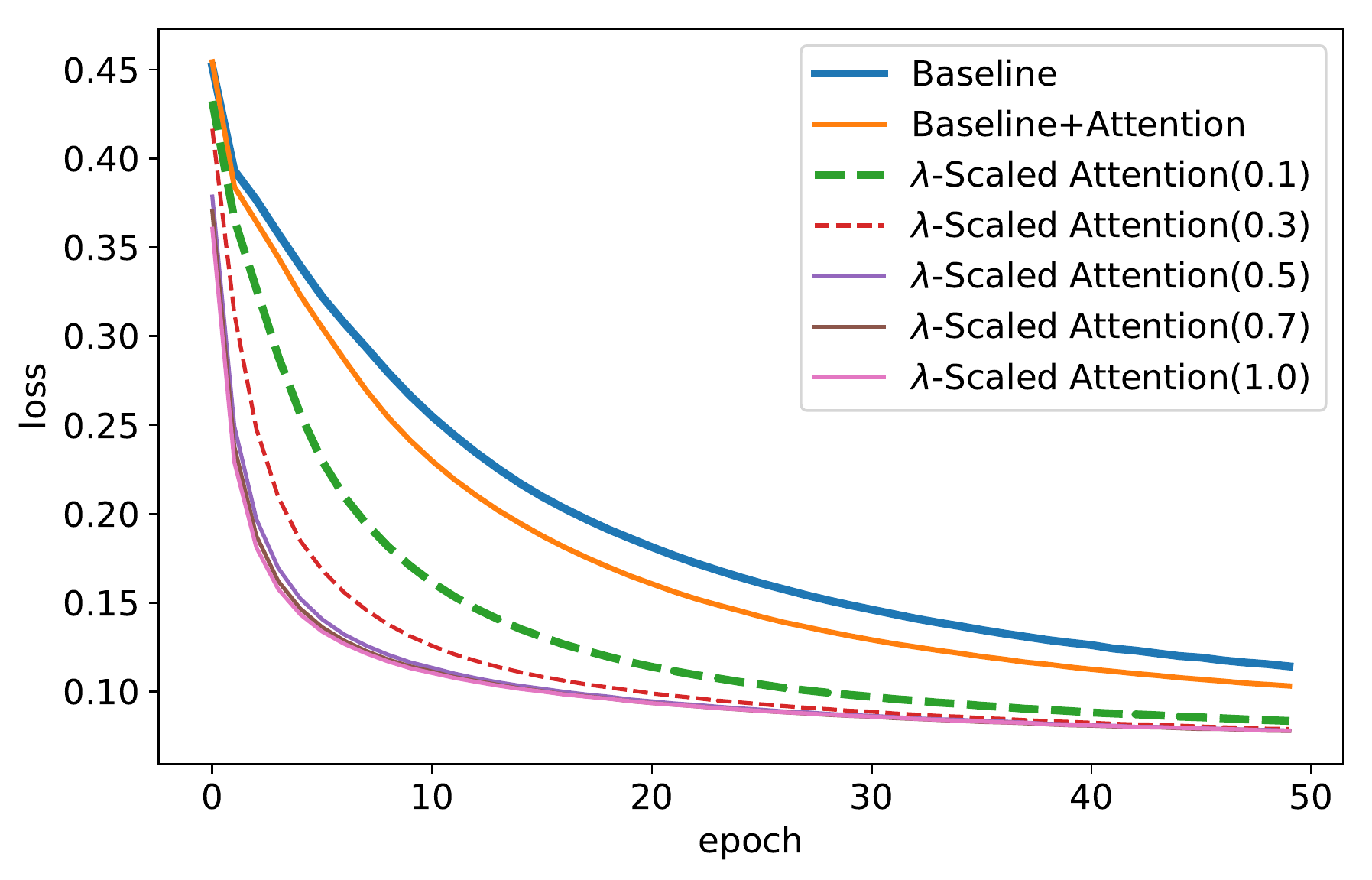}\label{bio_loss}}
	\hfil
	\subfloat[Validation Loss]
	{\includegraphics[width=9.1cm]
		{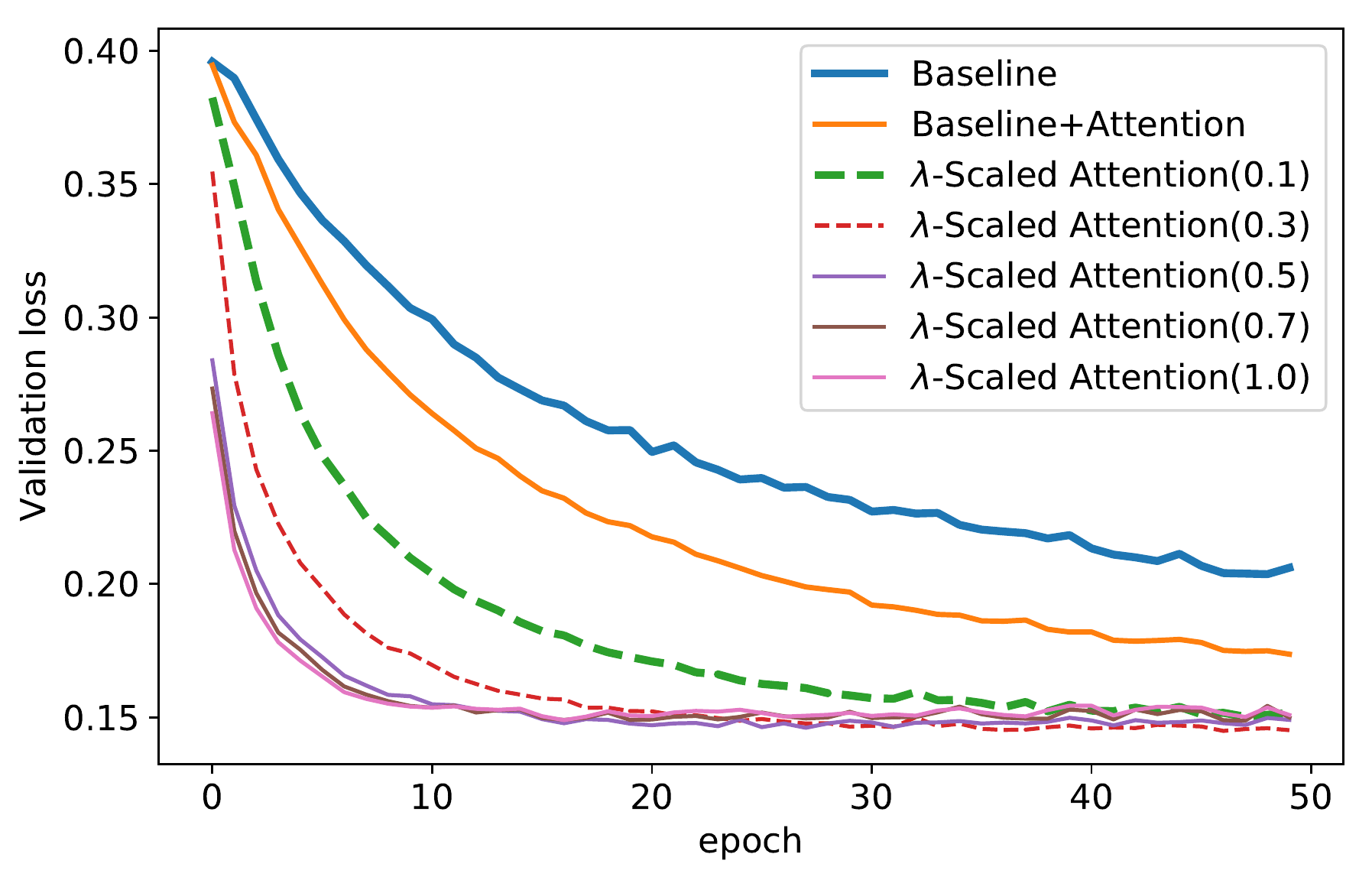}\label{bio_valLoss}}
	\caption{Biological Process: LSTM based training and validation loss curve.}
	\label{LossBIO}
\end{figure*}

\subsection{$\lambda$-\textit{scaled attention network} performance analysis}
In this section, we comprehensively evaluate the effect of the proposed $\lambda$-\textit{scaled attention network} onto the function prediction task by comparing it with the two baseline models used at the sub-sequence level. The baseline models are as follows:
\begin{enumerate}
	\item \textit{Baseline classifier}: This model does not use \textit{attention} technique. Specifically, it only uses the last hidden state from the deep recurrent processing layer, denoted $h_T$, as the intermediate representation for the input protein segment.
	
	\item \textit{Baseline + attention-based classifier}: This model includes the standard attention technique proposed by Bahdanau et. al. \cite{Attention}. Intermediate representation for the input protein segment is the aggregation of the weighted hidden states.
\end{enumerate}

There are two aspects to evaluation: (i) the performance evaluation based on the overall F1-scores, and (ii) the evaluation of the training process with respect to learning and convergence rate. Learning is the evaluation of the validation loss with respect to training loss. The lesser is the gap between the training loss and validation loss, the more effective learning is, and vice versa.

Moreover, we also evaluated the proposed $\lambda$-\textit{scaled attention network} by varying the embedding dimensions for the protein words (discussed in Section \ref{embd_result}). Note that the final classification model (i.e., an MLP network), already discussed in Section \ref{class_model}, is the same in all the experiments. 

\subsubsection{Superior overall performances}
Here, overall performances (with respect to the F1-score) of the proposed $\lambda$-\textit{scaled attention network} are evaluated against both the baseline models. Note that each of GRU and LSTM networks is explored as the deep recurrent processing layer in the above classifiers. The comparison results are shown in Tables \ref{LSTM_atten} (with LSTM unit) and \ref{GRU_atten} (with GRU unit) for both the BP and MF datasets.

The experiments using the LSTM as the deep recurrent processing layer (Table \ref{LSTM_atten}) demonstrate that the proposed $\lambda$-\textit{scaled attention network} based classifiers (with $\lambda$ = 1.0) are superior to both the baseline models. For the BP dataset, with segment sizes 80, 100, and 120, the improvements recorded over the \textit{baseline classifiers} were +3.12\%, +4.63\%, and +5.64\%, respectively. The corresponding improvements over the \textit{baseline+attention-based classifiers} were +1.26\%, +2.01\%, and +1.74\%, respectively.

Similarly, with the MF dataset (Table \ref{LSTM_atten}), the observed improvements are as follows: compared to the \textit{baseline classifiers} and with segment sizes 80, 100, and 120, improvements of +3.76\%, +5.51\%, and +7.54\% were recorded, respectively. The corresponding improvements with respect to the \textit{baseline+attention-based classifiers} are +2.48\%, +4.67\%, and +3.87\%, respectively. Moreover, the above observations are also true for all the $\lambda$ values with the proposed $\lambda$-\textit{scaled attention network} based classifiers.

Further experiments using the GRU as the deep recurrent processing layer (Table \ref{GRU_atten}) also suggest the superior results with the proposed $\lambda$-\textit{scaled attention network} based classifiers. These improvements hold irrespective of the $\lambda$ value.

\subsubsection{Efficient learning and faster convergence}\label{fastnefficient}
\begin{figure*}[!t]
	\centering
	\subfloat[Training Loss]
	{\includegraphics[width=9.1cm]
		{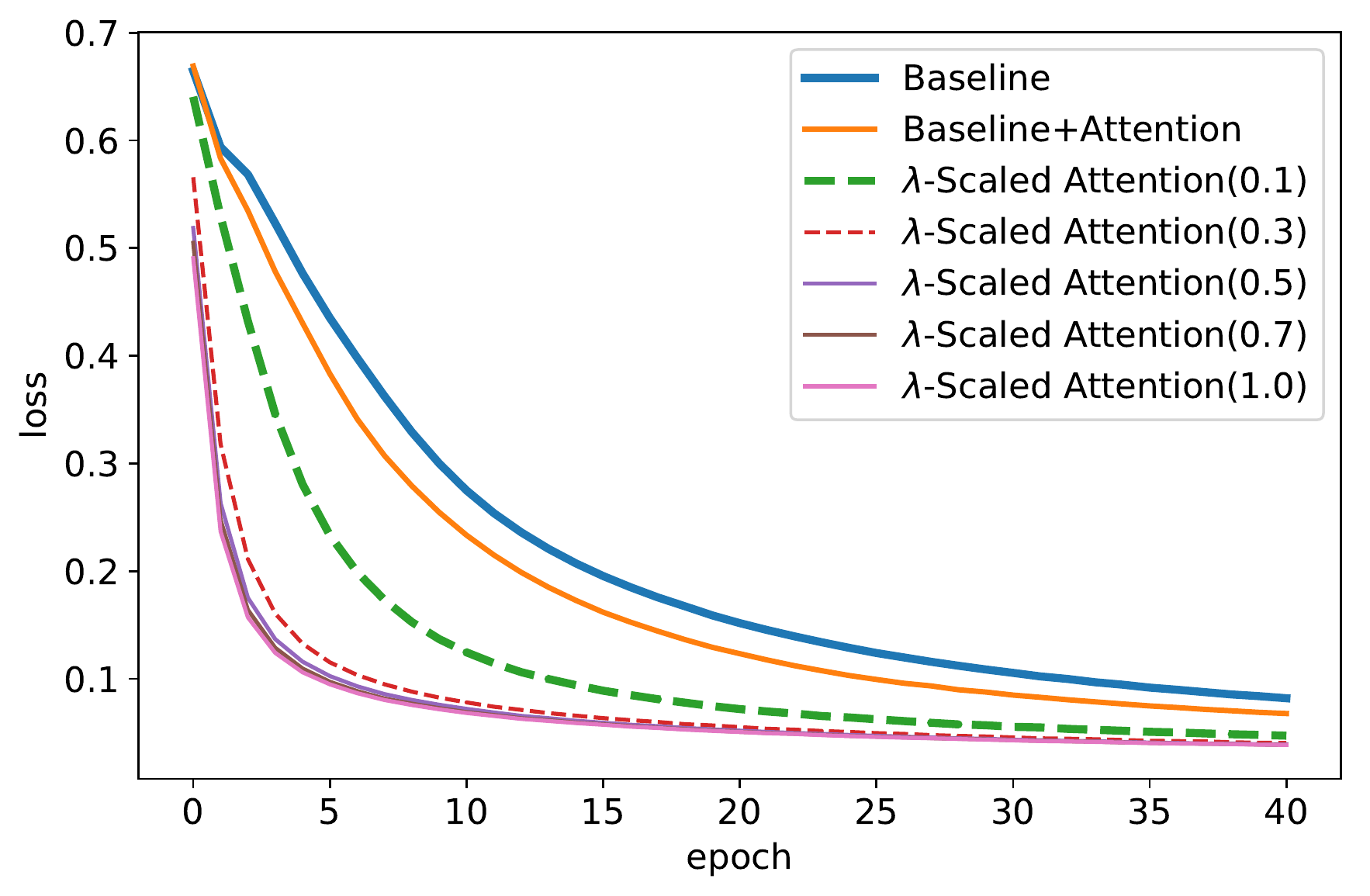}\label{mol_loss}}
	\hfil
	\subfloat[Validation Loss]
	{\includegraphics[width=9.1cm]
		{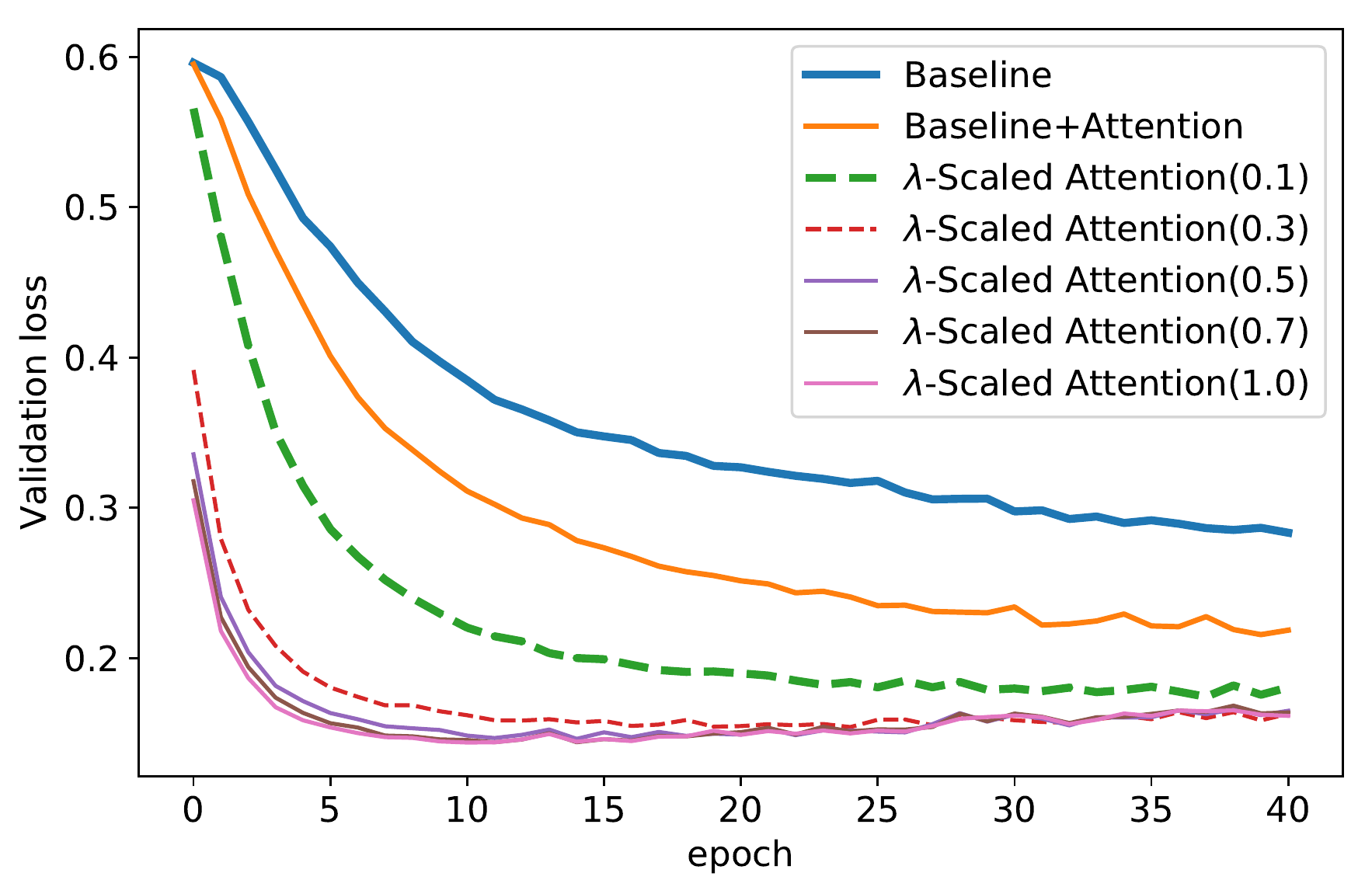}\label{mol_valLoss}}
	\caption{Molecular Function: LSTM based training and validation loss curve.}
	\label{LossMOL}
\end{figure*}

The curves for the training and validation losses is used to study the effect of the proposed $\lambda$-\textit{scaled attention technique} on the training efficiency and the convergence of the sub-sequence classifier. For a comparative analysis, the curves showing the training and validation losses of different classifiers are shown in Figures \ref{LossBIO} (BP) and \ref{LossMOL} (MF).

A high gap between the training and validation loss curves for the \textit{baseline classifiers} (Figures \ref{LossBIO} (BP) and \ref{LossMOL} (MF)) demonstrate a poor learning. The learning, however, is improved for the \textit{baseline+attention-based classifiers}, but losses are still high. A significant reduction in the losses are observed for the proposed $\lambda$-\textit{scaled attention network} based classifiers, demonstrating the best learning among all.

Moreover, Figures \ref{LossBIO} (BP) and \ref{LossMOL} (MF) also highlight the faster convergence of the proposed $\lambda$-\textit{scaled attention network} based classifier, when compared to both the \textit{baseline} and \textit{baseline+attention based classifiers}. The convergence with the proposed $\lambda$-\textit{scaled attention network} takes roughly half the amount of time the other baseline methods are taking. The convergence of the classifiers under the proposed $\lambda$-\textit{scaled attention network}, however, is delayed as the $\lambda$ value is decremented.

\begin{table*}[t]
	\caption{Classification Accuracy for the  $\lambda$-\textit{scaled attention network} based classifiers with $\lambda = 1$ ($d$ = embedding dimension)}\label{embd_atten}
	\centering
	\begin{tabular}{|p{1.1cm}|p{1.1cm}||p{0.87cm}|p{0.87cm}|p{0.87cm}|p{0.87cm}||p{0.87cm}|p{0.87cm}|p{0.87cm}|p{0.87cm}|}
		\hline
		\multicolumn{2}{|c||}{ }	& \multicolumn{4}{c||}{LSTM} & \multicolumn{4}{c|}{GRU}\Tstrut\\ [0.4ex]
		
		\hline \Tstrut
		Dataset & Seg. Size & \textit{d} = 32 & \textit{d} = 48 & \textit{d} = 64 & \textit{d} = 80 & \textit{d} = 32 & \textit{d} = 48 & \textit{d} = 64 & \textit{d} = 80\Tstrut\\
		\hline
		
		\textit{Biological} & 80 & 56.74 & 58.93 & 59.73 & \textbf{60.17} & 57.43 & 59.32 & 59.69 & \textbf{59.96}\Tstrut\\ [0.4ex]
		\textit{Process} & 100 & 57.26 & 58.69 & 59.88 & \textbf{60.09} & 57.28 & 58.66 & 59.35 & \textbf{59.51}\Tstrut\\ [0.4ex]
		& 120 & 57.09 & 58.87 & 59.75 & \textbf{59.91} & 57.34 & 58.57 & 59.18 & \textbf{59.68}\Tstrut\\ [0.4ex] \hline

		\textit{Molecular} & 80 & 69.73 & 71.58 & 71.61 & 71.59 & 69.99 & 70.63 & 71.53 & \textbf{71.39}\Tstrut\\ [0.4ex]
		\textit{Function} & 100 & 70.11 & 71.09 & 71.15 & \textbf{71.57} & 70.22 & 70.79 & 71.21 & \textbf{71.25}\Tstrut\\ [0.4ex]
		& 120 & 70.08 & 71.12 & 70.93 & 71.15 & 69.69 & 70.38 & 70.72 & \textbf{70.84}\Tstrut\\ [0.4ex] \hline
	\end{tabular}
\end{table*}

\begin{table*}[t]
	\caption{Overall Classification Accuracy (* mark indicates proposed methods)}
	\label{overall_acc}
	\centering
	\begin{tabular}{|p{0.45cm}|p{3.9cm}|p{1.75cm}||p{1.17cm}|p{1.17cm}|p{1.21cm}||p{1.17cm}|p{1.17cm}|p{1.21cm}|}\hline 
		& \multicolumn{2}{c||}{} & \multicolumn{3}{c||}{\textit{Biological Process} (\%)} & \multicolumn{3}{c|}{\textit{Molecular Function} (\%)}\Tstrut\\ [0.4ex]
		\hline \Tstrut
		S. No. & Approach & Recurrent cell /embedding dimension & Average Recall & Average Precision & Average F1-Score & Average Recall & Average Precision & Average F1-Score\Tstrut\\
		\hline
		
		1 & MLDA \cite{MLDA_protFun} & -- & 49.42 & 52.61 & 49.27 & 58.29 & 60.20 & 57.91\Tstrut\\ [0.4ex]
		\hline
		
		2 & ProtVecGen-100 \cite{Paper_mine} & LSTM/32 & 53.15 & 54.42 & 52.11 & 63.93 & 65.25 & 63.39\Tstrut\\ [0.4ex] 
		\hline
		
		3 & ProtVecGen-Plus \cite{Paper_mine} & LSTM/32 & 56.42 & 56.65 & 54.65 & 66.93 & 67.42 & 65.91\Tstrut\\ [0.4ex] \hline
		
		4 & ProtVecGen-Plus + MLDA \cite{Paper_mine} & -- & 58.19 & 58.80 & 56.68 & 68.62 & 68.27 & 67.12\Tstrut\\ [0.4ex] \hline
		
		5 & Proposed-100* & LSTM/32 & 58.35 & 59.96 & 57.26 & 70.86 & 71.95 & 70.11\Tstrut\\ [0.4ex] \hline
		
		6 & Proposed-100* & GRU/32 & 58.54 & 59.95 & 57.28 & 71.01 & 72.12 & 70.22\Tstrut\\ [0.4ex] \hline
		
		7 & Proposed-100* & LSTM/64 & 60.59 & 63.17 & 59.88 & 71.85 & 73.07 & 71.15\Tstrut\\ [0.4ex] \hline
		
		8 & Proposed-100* & GRU/64 & 59.95 & 62.71 & 59.35 & 71.95 & 73.06 & 71.21\Tstrut\\ [0.4ex] \hline
	\end{tabular}
\end{table*}

\subsubsection{With respect to parameter embedding dimension}\label{embd_result}
Word embedding creates dense representation for the protein words, where each dimension represents a hidden attribute crucial to the meaning of the word. A higher dimension is often advantageous creating a more meaningful representations for the word, however, the increase in the dimension also causes a significant increase in the number of trainable parameters. Here, the proposed $\lambda$-\textit{scaled attention network} based classifier is evaluated by varying the embedding dimension of the protein words. While the default embedding dimension (\textit{d}) is $d = 32$, the proposed $\lambda$-\textit{scaled attention network} based classifier (with $\lambda = 1.0$) is also evaluated for the other values of the $d = [48, 64, 80]$. The results are shown in Table \ref{embd_atten} with both GRU and LSTM as the  deep recurrent layer. The number of trainable parameters when $d$ = 32 is $\approx$ 5.20 million.

As shown in Table \ref{embd_atten}, the average F1-score increases significantly upon increasing the embedding dimension of the protein words from $d$ = 32 to $d$ = 48. The number of trainable parameters when $d$ = 48 is $\approx$ 7.76 million. With LSTM, for the biological process dataset, the margins of improvement are +2.19\%, +1.43\%, and +1.78\% with segment sizes 80, 100, and 120, respectively. The corresponding improvements for the molecular function dataset are +1.85\%, +0.98\%, and +1.04\%, respectively. Similar performance enhancements were  observed with the GRU as the deep recurrent layer. Further increasing the embedding dimension (i.e., when $d$ = 64) only results in small improvements in the average F1-score, while a marginal/no improvement is observed by increasing the embedding dimension up to $d$ = 80. 

We, therefore, choose to stick to the results obtained using $d$ = 64 (with $\approx$ 10.33 million trainable parameters), which has significantly less trainable parameters than $d$ = 80 (with $\approx$ 12.90 million parameters).

\subsection{Performance comparison with work from the literature}
In Table \ref{overall_acc} we have listed performances of the proposed method, denoted as proposed-100* (LSTM/32), along with the existing work from the literature for protein function prediction, where `LSTM' and `32' indicates recurrent cell and embedding dimension, respectively. The work from the literature includes: (i) MLDA approach \cite{MLDA_protFun}, which is based on the \textit{tf-idf}, (ii) ProtVecGen-100 \cite{Paper_mine}, which is a single-segment based approach, (iii) state-of-the-art ProtVecGen-Plus \cite{Paper_mine}, which is a multi-segment based approach, and (iv) ProtVecGen-Plus+MLDA \cite{Paper_mine}, which is a hybrid approach.

\begin{figure*}[!t]
	\centering
	\subfloat[Biological Process]
	{\includegraphics[width=12.0cm]
		{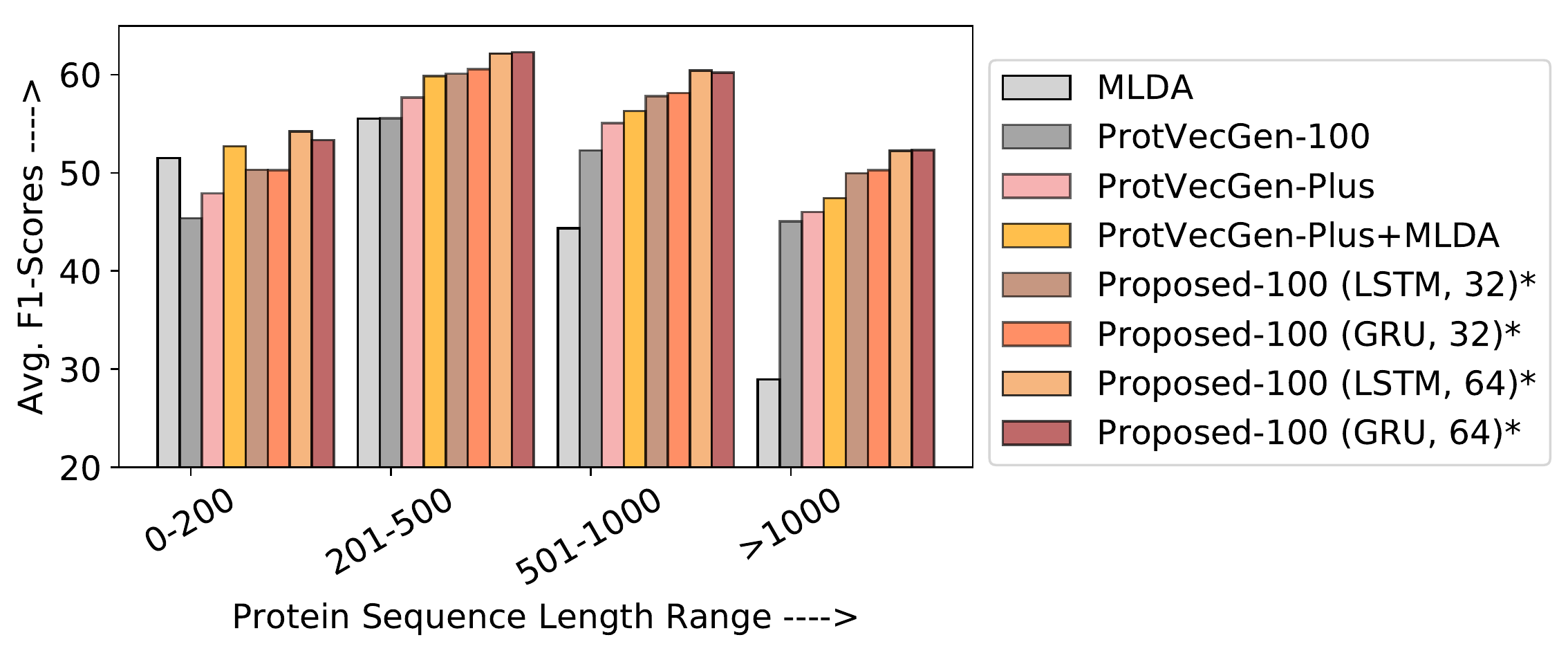}\label{bio_len_wise}}
	\hfil
	\subfloat[Molecular Function]
	{\includegraphics[width=12.0cm]
		{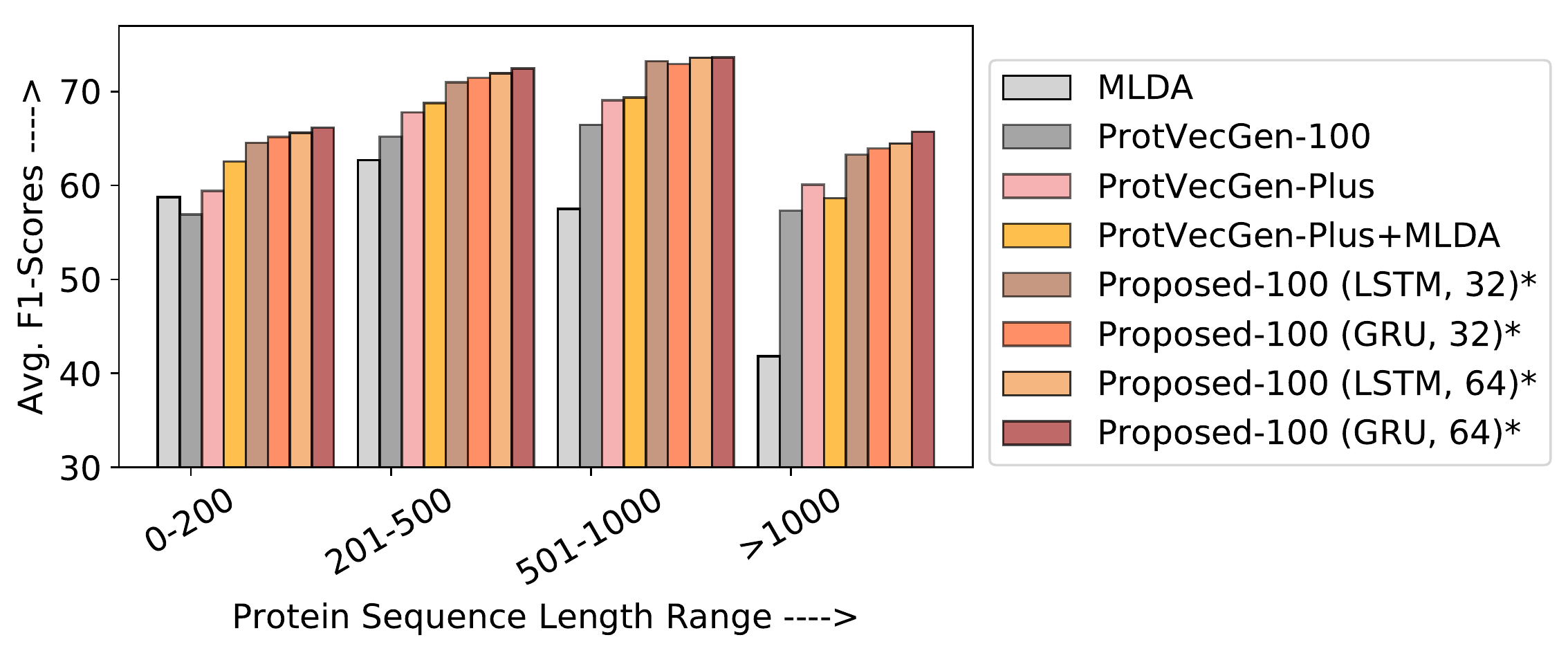}\label{mol_len_wise}}
	\caption{Length-wise Evaluation: For the proposed methods, the configurations are denoted in brackets.}
	\label{len_wise}
\end{figure*}

Additionally, their performances were also evaluated with respect to their ability to deal with the protein sequences of different lengths. This is done by splitting the test datasets into the following groups based on their sequence lengths: G1: (0 – 200), G2: (201 – 500), G3: (501 – 1000), and G4: $>$ 1000. The range for each group specifies the permissible number of amino acid residues in a sequence. These are shown in Figures \ref{bio_len_wise} (BP) and \ref{mol_len_wise} (MF).

As shown in Table \ref{overall_acc}, compared to the classifier trained on the MLDA features \cite{MLDA_protFun}, the classifiers trained with the proposed-100* (LSTM/32) features have improved significantly: the margins of improvement in the average F1-scores are +7.99\% for the BP dataset and +12.20\% for the MF dataset. The proposed-100* (LSTM/32) features  also showed huge improvements when validated for the protein sequences in the groups G2, G3, and G4 (See Figures \ref{bio_len_wise} (BP) and \ref{mol_len_wise} (MF)).

In addition, the proposed-100* (LSTM/32) features effectively outperformed both the single-segment based ProtVecGen-100 \cite{Paper_mine} and a multi-segment based ProtVecGen-Plus \cite{Paper_mine} features with the respective improvements of +5.15\% and +2.61\% for the BP dataset. The corresponding improvements for the MF dataset are +6.72\% and +4.20\%, respectively. Besides this, the evaluation of the proposed-100* (LSTM/32) on protein sequences of different lengths, as shown in Figures \ref{bio_len_wise}  (BP) and \ref{mol_len_wise} (MF), clearly establishes the superiority of the proposed features over them.

Finally, comparison with the hybrid ProtVecGen-Plus+MLDA \cite{Paper_mine} approach also shows better results for the proposed-100* (LSTM/32) features by margins of +0.58\% (for BP) and +2.99\% (for MF). The recall and precision values also exhibit similar behavior. Furthermore, the proposed-100* (LSTM/32) based classifier has again proven to be superior when validated against protein sequences of different lengths, as shown in Figures \ref{bio_len_wise} (BP) and \ref{mol_len_wise} (MF). The proposed-100* (GRU/32) features also exhibit similar improved performances. The best overall results, however, are observed with the proposed-100* (LSTM/64) and proposed-100* (GRU/62) features.

\section{Conclusion}\label{conclusion}
In the paper, we explored the attention technique for the protein sequences that help in recognizing key protein words from the sequences. These keywords are useful in converting sequences to a meaningful representation for the classification task. However, in general, the standard attention technique suffers due to the weak semantics of the protein words. This poses a difficulty for the standard attention technique to construct a good representation for protein sequences. The novel challenges faced by the standard attention technique addressed within this work include (i) vanishing attention score problem and (ii) high variations in the attention distribution.

In order to deal with the above challenges, we introduced a novel $\lambda$-\textit{scaled attention} technique to find a good representation of protein sequences, where $\lambda$ is a parameter that assists in controlling the training process. The proposed attention technique helps not only in solving the vanishing attention score problem, but also helps in reducing the high variations in the attention distribution. This has a significant effect towards making the training more efficient and faster, when compared to the standard attention technique. The proposed $\lambda$-\textit{scaled attention} technique is further used to build the $\lambda$-\textit{scaled attention network} for the protein function prediction task at the protein sub-sequence level.

Overall, for the protein function prediction task, the proposed $\lambda$-\textit{scaled attention} technique demonstrated a significant effect on the predictive performances outperforming other existing approaches by substantial margins. More importantly, performances across the protein sequences of different lengths are consistently maintained.

\ifCLASSOPTIONcaptionsoff
  \newpage
\fi


\vfill


\begin{thebibliography}{1}
\bibitem{spike_protein1}
Li, F., 2016. ``Structure, function, and evolution of coronavirus spike proteins". \emph{Annual review of virology}, 3, pp.237-261.

\bibitem{spike_protein2}
Walls, A., Park, Y.J., Tortorici, M.A., Wall, A., Mcguire, A. and Veesler, D., 2020. ``Structure, Function, and Antigenicity of the SARS-CoV-2 Spike Glycoprotein". \emph{Cell}, 181(2), pp.281-292.

\bibitem{bacteria}
Durmus Tekir, S., Cakir, T. and Ulgen, K., 2012.  ``Infection strategies of bacterial and viral pathogens through pathogen–human protein–protein interactions". \emph{Frontiers in Microbiology}, 3, p.46.
	
\bibitem{lang_life}
Heinzinger, M., Elnaggar, A., Wang, Y., Dallago, C., Nechaev, D., Matthes, F. and Rost, B., 2019. ``Modeling aspects of the language of life through transfer-learning protein sequences". \emph{BMC bioinformatics}, 20(1), p.723.

\bibitem{NextGenSeq}
Metzker, Michael L. ``Sequencing technologies—the next generation." \emph{Nature Reviews Genetics} 11, no. 1 (2010): 31.

\bibitem{FANNGO}
Clark, Wyatt T., and Predrag Radivojac.``Analysis of protein function and its prediction from amino acid sequence." \emph{Proteins: Structure, Function, and Bioinformatics} 79.7 (2011): 2086-2096.

\bibitem{MLDA_protFun}
Wang, H., Yan, L., Huang, H., \& Ding, C. (2017). ``From Protein Sequence to Protein Function via Multi-Label Linear Discriminant Analysis". \emph{IEEE/ACM Transactions on Computational Biology and Bioinformatics (TCBB)}, 14(3), 503-513.

\bibitem{Paper_mine}
Ranjan, A., Fahad, M.S., Fernandez-Baca, D., Deepak, A. and Tripathi, S., 2019. ``Deep Robust Framework for Protein Function Prediction using Variable-Length Protein Sequences''. \emph{IEEE/ACM transactions on computational biology and bioinformatics}. doi: 10.1109/TCBB.2019.2911609.

\bibitem{DNA_bind}
Hu, S., Ma, R. and Wang, H., 2019. ``An improved deep learning method for predicting DNA-binding proteins based on contextual features in amino acid sequences". \emph{PloS one}, 14(11), p.e0225317.

\bibitem{anticancer}
Yi, H.C., You, Z.H., Zhou, X., Cheng, L., Li, X., Jiang, T.H. and Chen, Z.H., 2019. ``ACP-DL: a deep learning long short-term memory model to predict anticancer peptides using high-efficiency feature representation". \emph{Molecular Therapy-Nucleic Acids}, 17, pp.1-9.

\bibitem{prolango}
Cao, R., Freitas, C., Chan, L., Sun, M., Jiang, H. and Chen, Z., 2017. ``ProLanGO: protein function prediction using neural machine translation based on a recurrent neural network". \emph{Molecules}, 22(10), p.1732.

\bibitem{Seq_motif}
Ben-Hur, A. and Brutlag, D., 2006. ``Sequence motifs: highly predictive features of protein function". In \emph{Feature Extraction} . Springer, Berlin, Heidelberg, pp. 625-645.
	
\bibitem{Attention}
Bahdanau, D., Cho, K. and Bengio, Y., 2014. ``Neural machine translation by jointly learning to align and translate". \emph{arXiv preprint arXiv}:1409.0473.

\bibitem{hierarchical_attention}
Yang, Z., Yang, D., Dyer, C., He, X., Smola, A. and Hovy, E., 2016, June. ``Hierarchical attention networks for document classification". \emph{In Proceedings of the 2016 conference of the North American chapter of the association for computational linguistics: human language technologies}, pp. 1480-1489.

\bibitem{attention_sentiment1}
Chen, H., Sun, M., Tu, C., Lin, Y. and Liu, Z., 2016, November. ``Neural sentiment classification with user and product attention." \emph{In Proceedings of the 2016 conference on empirical methods in natural language processing} (pp. 1650-1659).

\bibitem{attention_sentiment2}
Zhou, X., Wan, X. and Xiao, J., 2016, November. Attention-based LSTM network for cross-lingual sentiment classification. In Proceedings of the 2016 conference on empirical methods in natural language processing (pp. 247-256).

\bibitem{dropout}
Srivastava, N., Hinton, G., Krizhevsky, A., Sutskever, I. and Salakhutdinov, R., 2014. ``Dropout: a simple way to prevent neural networks from overfitting." \emph{The Journal of Machine Learning Research}, 15(1), pp.1929-1958.

\bibitem{GRU}
Cho, K., van Merriënboer, B., Gulcehre, C., Bahdanau, D., Bougares, F., Schwenk, H. and Bengio, Y., 2014, October. ``Learning Phrase Representations using RNN Encoder–Decoder for Statistical Machine Translation". \emph{In Proceedings of the 2014 Conference on Empirical Methods in Natural Language Processing (EMNLP)} (pp. 1724-1734).

\bibitem{LSTM}
Hochreiter, S. and Schmidhuber, J., 1997.``Long short-term memory". \emph{Neural computation}, 9(8), pp.1735-1780.	

\bibitem{BiLstm}
Graves, A. and Schmidhuber, J., 2005. ``Framewise phoneme classification with bidirectional LSTM and other neural network architectures." \emph{Neural Networks}, 18(5-6), pp.602-610.

\bibitem{adam}
Kingma, D.P. and Ba, J., 2014. ``Adam: A method for stochastic optimization". \emph{arXiv preprint arXiv:1412.6980}.

\bibitem{uniprot}
UniProt Consortium, 2014. ``UniProt: a hub for protein information". \emph{Nucleic Acids Research}, 43(D1), pp.D204-D212.

\bibitem{GO}
Ashburner, M., Ball, C.A., Blake, J.A., Botstein, D., Butler, H., Cherry, J.M., Davis, A.P., Dolinski, K., Dwight, S.S., Eppig, J.T. and Harris, M.A., 2000. ``Gene Ontology: tool for the unification of biology". \emph{Nature Genetics}, 25(1), p.25.

\bibitem{Metrices1}
Sorower, M.S., 2010. ``A literature survey on algorithms for multi-label learning''. \emph{Oregon State University, Corvallis}, 18, pp.1-25.

\bibitem{Metrices2}
Zhang, M.L. and Zhou, Z.H., 2014. ``A review on multi-label learning algorithms". \emph{IEEE transactions on Knowledge and Data Engineering}, 26(8), pp.1819-1837.
\end{thebibliography}
\end{document}